\documentclass{article}

\usepackage[preprint, nonatbib]{neurips_2021}

\usepackage[T1]{fontenc}
\usepackage[utf8]{inputenc}

\usepackage{microtype}

\usepackage[numbers]{natbib}

\usepackage{xurl, hyperref}
\usepackage[super]{nth}
\usepackage[ruled]{algorithm2e}
\usepackage{float, graphicx, wrapfig, subcaption}
\usepackage{booktabs, tabularx, multirow, tablefootnote}
\usepackage{listings}
\usepackage{nicefrac}
\usepackage{amsmath, amsthm}

\title{Generalization over different cellular automata rules learned by a deep feed-forward neural network}

\author{%
 Marcel Aach$^{1,2}$, Jens Henrik Goebbert$^1$, Jenia Jitsev$^{1,3}$ \\
 $^1$ Jülich Supercomputing Center (JSC), Forschungszentrum Jülich, \\ 
 52425 Jülich, Germany \\ 
 $^2$ University of Cologne, Mathematical Institute, 50931 Cologne, Germany \\
 $^3$ Helmholtz AI, Forschungszentrum Jülich, 52425 Jülich, Germany \\
 \texttt{\{m.aach, j.goebbert, j.jitsev\}@fz-juelich.de}
}

\begin{document}


\maketitle

\begin{abstract}
To test generalization ability of a class of deep neural networks, we randomly generate a large number of different rule sets for 2-D cellular automata (CA), based on John Conway’s Game of Life. Using these rules, we compute several trajectories for each CA instance. A deep convolutional encoder-decoder network with short and long range skip connections is trained on various generated CA trajectories to predict the next CA state given its previous states. Results show that the network is able to learn the rules of various, complex cellular automata and generalize to unseen configurations. To some extent, the network shows generalization to rule sets and neighborhood sizes that were not seen during the training at all. \footnote{Code to reproduce the experiments is publicly available at: \url{https://github.com/SLAMPAI/generalization-cellular-automata}}

\end{abstract}


\section{Introduction}
Deep learning has become a powerful tool to address broad range of problems in various scientific and technological disciplines by training task-solving models on very large datasets collected in different domains. Training on a dataset under certain conditions always raises the fundamental question of generalizability of the obtained model when applied to the problem under different conditions or on entirely different data than the one utilized during the training. The ability to generalize well across conditions, datasets and tasks is the ultimate goal of training any machine learning model, and deep learning methods are here no exception. 

Deep learning models trained on large amount of data also do show lack of generalization when facing varying conditions, indicating that often the solutions obtained after training are narrowly tailored and do not capture the full extent of the problem structure to be solved \cite{Recht2019, Hendrycks2019}. The motivation of this paper is to study generalization on a problem setting of learning the rules underlying pattern generation in cellular automata (CA) \cite{Neumann1966, Ulam1962}. CA are defined by rather simple, spatially local rules that instantiate discrete dynamical systems which may exhibit very complex behavior, producing chaotic and hardly unpredictable patterns \cite{Berlekamp2004, Wolfram2002}. 

These properties makes CA a very good candidate for examining generalization capability of a deep learning method: given a set of ground truth CA rules, we can generate trajectories corresponding to each CA instance, use those trajectories for training a network model on a sufficiently complex dataset and then test the trained model on sets of rules that were either already used during the training or on sets of rules that the network has not yet seen before. The ability to quickly generate complex CA datasets with given rule sets allows us thus to systematically test to what extent a deep learning method can handle both already seen and unseen conditions, thus assessing its capability to generalize.

For the deep neural network model architecture we follow the recent advancements in the field of image recognition and segmentation, using an encoder-decoder style convolutional neural network (CNN) with elements from the SegNet \cite{Badrinarayanan2016}, ResNet \cite{He2016} and U-Net \cite{Ronneberger2015}. To evaluate model's ability to generalize we use two-dimensional CA. As input data, network model will take four consecutive time steps of the CA and predict on the output the fifth time step.

One of the most famous CA is the classical Game of Life (GoL) by John Conway \cite{Gardner1970} which serves as a starting point in this paper. It consists of an array of cells in space that are connected to each other. Each cell can take the value zero or one which correspond to it being either \textbf{dead} or \textbf{alive}. Starting from an initial configuration, each one of the cells changes its state in the next iteration based on its current state and the number of its surrounding (living or non-living) neighbors. A dead cell comes to life if it has exactly three neighbors that are alive (reproduction), a living cell with two or three neighbors that are alive stays alive (survival) and if a living cell has less than two neighbors that are alive or more than three, it dies (underpopulation and overpopulation). The total number of neighbors a cell has is defined by the neighborhood size, which for the Game of Life is the $3 \times 3$ Moore neighborhood. We modify these original rules of the GoL to get several hundred different rule sets which then are used to generate the training, validation and test data sets for the neural network. 

Rules used in the Game of Life and in other comparably simple cellular automata may be employed to describe processes that govern real world's physics, as despite their apparent simplistic form they can be shown to be Turing complete and thus represent any computable function \cite{Berlekamp2004, Wolfram2002, wolfram2020project}. Konrad Zuse \cite{Zuse91} brought up the possibility that the whole universe itself is indeed a digital CA and that the complex laws of physics we can observe are just derived from a simple, local set of rules this CA follows. This would mean that space and time in our universe are based on a discrete grid of cells where a cell is the smallest possible unit that exists. In contrast to classical physics that assumes all physical quantities to be continuous, the field of quantum physics already introduced the idea that some quantities can only take discrete values. As Wolfram \cite{Wolfram2002} points out, such a discrete model would actually be better in explaining and capturing the complexity that arises in the real world than numerical methods that rely on some sense of smoothness. There also exist continuous variants of CA that operate on continuous state variables, based on locally defined interaction kernels~\cite{Wolfram2002}. Although so far these are just theoretical considerations, CA models have already successfully been applied to explain physical phenomena such as gas and fluid dynamics \cite{hpp73}.  

\section{Related Work}

\subsection{Convolutional Neural Networks for Image Processing}
Generating very deep neural networks by stacking many convolutional layers on top of each other has been the main strategy in recent years to improve accuracy for image recognition tasks. It has however been shown, that simply stacking layers on top of each other does not work in straightforward way. Without additional measures, the performance actually drops with the addition of more layers (degradation problem) \cite{He2016}. Further increasing depth of the network was strongly hinted to be the way to further increase model performance, such that the inability to stack the layers towards deeper architectures became problematic. To address this problem, deep residual networks (ResNet) were introduced. In a ResNet-architecture the output mapping of each layer $\mathcal{H}(x)$ is composed of the residual transformation of the original input $\mathcal{F}(x)$ and the original input itself, such that residual mapping $\mathcal{F}(x) = \mathcal{H}(x) - x$ has to be learned at each processing stage. The desired output mapping becomes $\mathcal{H}(x) = \mathcal{F}(x) + x$. In terms of the structure of the network, this is realized using additive \textit{shortcut skip connections}, which perform identity mapping. This architecture allows proper gradient flow through the whole depth of the network and encourages learning of the features that are not yet extracted in the previous stages. In the original paper \cite{He2016} the authors use 31 convolutional layers (with an increasing number of filters), two pooling layers, and a fully connected dense layer for the output at the end.

ResNet architectures excels in object recognition given an input image, able to signal a probability whether an image contains a particular object, for example a car or an apple. In other applications, e.g in biomedical imaging, the output is often not only a single class label for the whole input image but instead a label for each part or even pixel of the input. This type of problem is referred to as image segmentation. For such tasks, the output dimension usually has to be the same or similar as the input dimension.

The SegNet architecture is well suited for this type of task. It is capable of pixel-wise semantic segmentation using an encoder-decoder design. For these type of models, there is an encoder network in the first part, that takes the input and processes it through multiple layers. The processed input is projected into so called \textit{latent space}, which contains the representation of the current observation the network has built from the input data. If learning goes well, the latent space may be low dimensional set of variables that correspond to compact description of the observed data as learned by the network. Second part of the network constitutes the decoder. Decoder maps the representation of the data from the latent space either back to the original input data space or to a target space of similar dimension. In the case of the SegNet, the decoder is an inverse of the encoder: a convolutional layer becomes a deconvolutional layer, a pooling layer becomes an upsampling layer. Using the encoder and decoder in a single model makes end-to-end training possible using either supervised or unsupervised losses for latent and target spaces. 

U-Net is a version of SegNet employing long-range skip connections that deliver feature information from encoder layers directly to decoder by concatenation to the output of respective decoder layers. In this way, information compressed via encoder into the latent space can be combined together again with less compressed low level feature encoder information at the decoder level.

\subsection{Deep Learning applied to Cellular Automata}
The strength of convolutional neural networks in dealing with input data that has local spatial structure make them suitable candidates for application to dynamical systems defined by local interactions which CA are. Because neural networks can (in theory) represent any function \cite{Cybenko1989}, together with the inductive bias that respects local structure we would expect near-optimal results when approximating the rules of cellular automata. A study on how CNN represent CA rules was published by Gilpin \cite{Gilpin2019}. In general the structure of such a CNN consists of two processing stages: the first stage identifies the current CA state by counting the number of living and non-living cells in the neighborhood of a single cell while the second stage transforms this information to the correct next cell state given the rules. Gilpin demonstrates, that using just one convolutional layer with a kernel size of $3 \times 3$ and enough filters, followed by several $1 \times 1$ convolutional layers, is capable of predicting any binary cellular automaton with a Moore neighborhood of size $3 \times 3 $ with an accuracy of $100 \%$. This network is fully convolutional but does not include any downsampling steps, thus keeping the original input dimensions across all layers.

While these results are very promising, it has also just recently been shown by Springer and Kenyon \cite{Springer2020} that a network needs to be "overcomplete", meaning that it should have significantly more parameters than theoretically needed to learn the representation of the Game of Life for the prediction of one or several time steps. The authors use a minimal neural network, also made of $3 \times 3$  and $1 \times 1$ convolutional layers, while fixing the number of filters and therefore the number of parameters. Predicting the next three, four and five time steps completely failed, and predicting the next one or two time steps only succeeded if the network was $2-10$ times overcomplete. 

\section{Experiment Design} 

\subsection{Rule Modifications}

\begin{figure*}[!htb]
\centering
\begin{minipage}{.25\textwidth}
    \includegraphics[width=\textwidth]{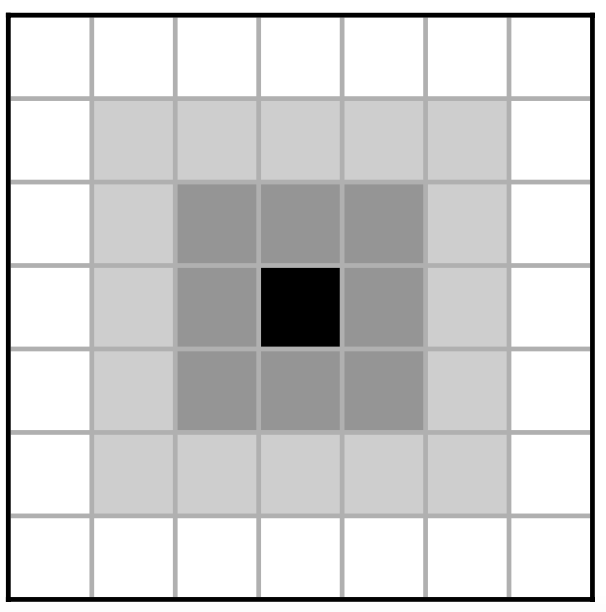}
    \subcaption{Moore nbhd.}
    \label{fig:extended_neighborhood}
    \includegraphics[width=\textwidth]{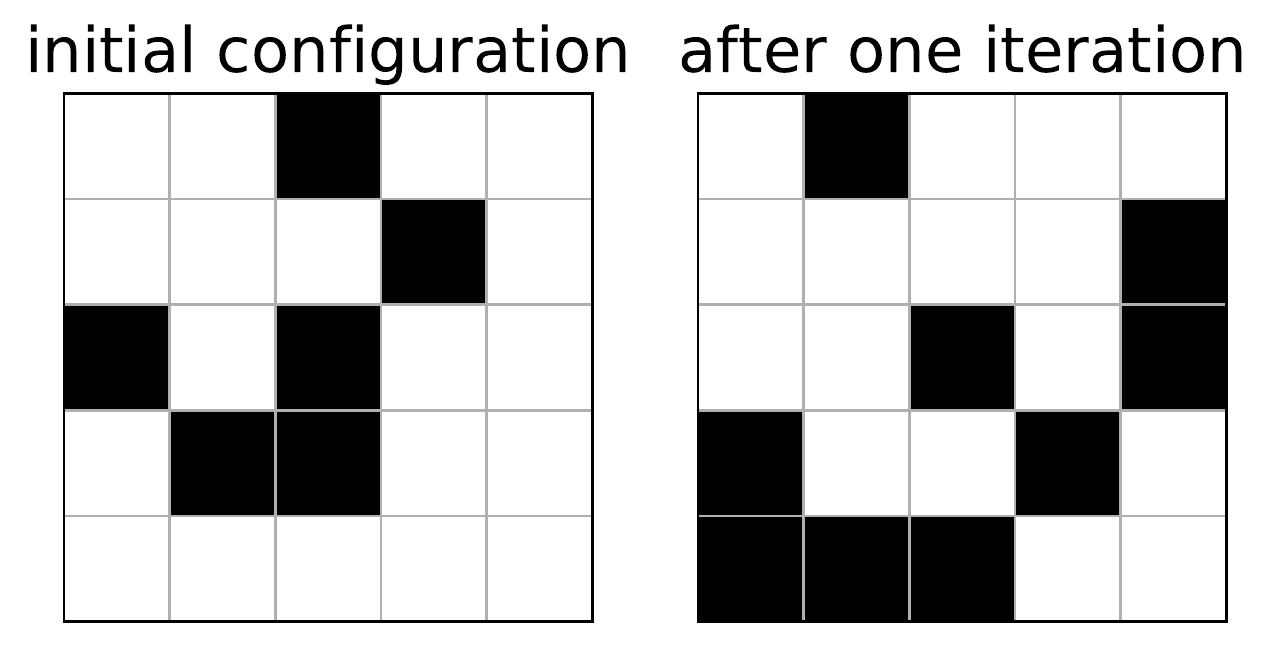}
    \subcaption{Cell transition}
    \label{fig:cell_transition}
\end{minipage}
\begin{minipage}{.7\textwidth}
    \includegraphics[width=\textwidth]{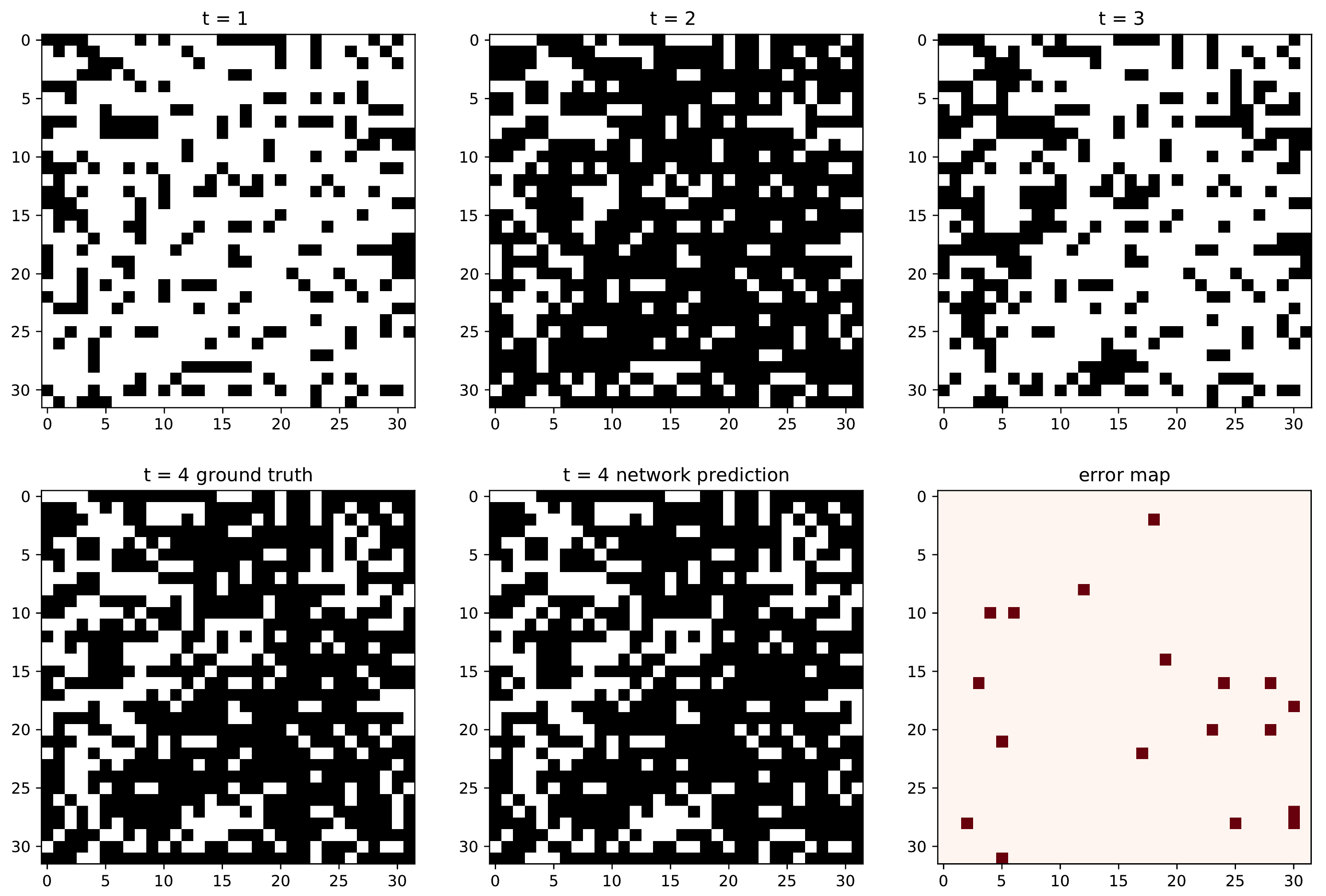}
    \subcaption{Trajectory of a CA}
    \label{fig:trajectory}
\end{minipage}
\caption{(a) The $3 \times 3$ Moore neighborhood of a single cell (black) is shown in dark grey and the $5 \times 5$ Moore neighborhood in light grey. (b) The rule Born: 1,4,7 Stay: 2,5,10,12 with the $5 \times 5$ Moore neighborhood and non-periodic boundary conditions visualized. The cell in the center survives, because it has exactly 5 living neighbors. (c) Trajectory of a (mostly) "alternating" rule, including the network prediction and error map for $t=4$. What was alive dies in the next time step and vice versa. $t=0$ is not shown.}
\end{figure*}

To fully assess the generalization performance of a neural network it is crucial to be able to control different aspects of the generated data and thus differences in datasets used for training and test. In the case of the Game of Life the simplest way to generate different data is to just change the initial starting configuration. But to achieve more challenging levels of difference that would also define different degrees of difficulty to generalize, we apply two modifications to the way the original GoL works:  
\begin{itemize}
    \item \textbf{Modification of the cell transition rules:} we change the number of living neighbors required for a cell to die, survive, or be born. This number is chosen randomly from a plausible set of numbers.
    \item \textbf{Modification of the cell neighborhood:} we also modify the size of the Moore neighborhood that influences the status of the cell, so the rule's spatial extent becomes larger, see figures \ref{fig:extended_neighborhood} and \ref{fig:cell_transition}. This also requires a new set of transition rules. 
\end{itemize}

The generation of rules is performed by randomly sampling integer numbers for the number of cells required for a cell to be born or to survive. We call this the Born/Stay notation \footnote{see \url{https://www.conwaylife.com/wiki/Cellular_automaton\#Rules}}, the classical Game of Life for example would be B3/S23. The numbers of course have to be plausible in the sense that for a cell neighborhood of size $3 \times 3$ the maximum of cells can be $8$ and the minimum should be at least $1$ to not immediately overpopulate the grid. Though one might expect that randomly sampling cell transition rules in this way would result in a large number of trivial rules (completely full or empty grids), this did not happen in practice. 

\subsection{Levels of Generalization}
When training and evaluating the networks performance we use a three step process: First we train the network using CA trajectories as input/output data. The input data are three consecutive time steps corresponding to states evolved from an initial configuration, stacked together. The output data to be predicted by the network is the next state. During the training process, we permanently validate the performance of the network on a validation data set, different from the training set in terms of initial configurations. This monitors overfitting. As a last step we load the weights of the best performing model from the validation step and test the model on a different set of data. The CA trajectories in the test set feature not only different initial configurations but might also arise from different cell transition rules using different neighborhoods sizes. 
\smallbreak

To check how well the network is able to generalize, we test for several levels of generalization. For each level we have a distinct data set (see also table \ref{generalization_datasets}):

\begin{itemize}
    \item \textbf{Simple generalization:} we train and validate the network on 300 different, randomly-generated transition rules of neighborhood size $3 \times 3$, then test on more unseen initial configurations (generated with the same rules). 
    
    \item \textbf{Level 1 generalization:} we train and validate the network on 300 different, randomly-generated transition rules of neighborhood sizes $3 \times 3$ and $5 \times 5$ and $7 \times 7$, then test on more unseen initial configurations (generated with the same rules).
    
    \item \textbf{Level 2 generalization:} we train and validate the network on 300 different, randomly-generated transition rules of neighborhood sizes $3 \times 3$ and $5 \times 5$ and $7 \times 7$, then test on more unseen initial configurations (generated with 30 \textit{different} randomly generated rules).
    
    \item \textbf{Level 3 generalization:} for the extrapolation task we train and validate the network on 300 different, randomly-generated transition rules of neighborhood sizes $3 \times 3$ and $5 \times 5$ and $7 \times 7$, then test on 30 unseen generated rules of neighborhood size $9 \times 9$. We also train a Level 3 interpolation version with neighborhood sizes  $3 \times 3$ and $5 \times 5$ and $9 \times 9$ in the training set and neighborhood size $7 \times 7$ in the test set.
    
\end{itemize}

\begin{table}
\begin{tabular}{c|cc}
                           & train neighborhood sizes & test neighborhood sizes              \\
                           \hline
Simple generalization      & 3x3                      & 3x3                                  \\
Level 1 generalization     & 3x3, 5x5, 7x7            & 3x3, 5x5, 7x7                        \\
Level 2 generalization     & 3x3, 5x5, 7x7            & 3x3, 5x5, 7x7 (different rules!) \\
Level 3 extrapolation     & 3x3, 5x5, 7x7            & 9x9                                  \\
Level 3 interpolation & 3x3, 5x5, 9x9            & 7x7                                  
\end{tabular}
\caption{Overview of the data sets used for the different levels of generalization.}
\label{generalization_datasets}
\end{table}

Overall we would expect very good results in the training and validation step for all levels of generalization. If the network can extract the rules, we would also expect very good test performance for the simple and level 1 generalization tasks. Since it has to adapt to a completely new set of rules in the test sets of the level 2 and level 3 generalization task, we expect poorer outcomes, but expect better results for the interpolation task than the extrapolation task. 

\subsection{A Hybrid Deep Residual U-Net Architecture}
The goal of our network is to extract the local cell transition rules from the observed image data and then apply them to the game board to get the next state of the CA. For this it makes sense to use the encoder-decoder structure (like in the SegNet and U-net). Using a binary crossentropy loss function, the job of the encoder part is to infer the rules and the current state of the CA from input observations and project those to the latent space. Then the decoder should use this information and generate the next state of the CA using latent space and encoder information provided via a long range skip connection. In contrast to the earlier mentioned work by Gilpin \cite{Gilpin2019} the encoder performs downsampling operations, changing the dimensions of the original image. The decoder however performs upsampling operations so at the end the original image dimensions are restored.

\begin{figure*}
    \includegraphics[width=\textwidth]{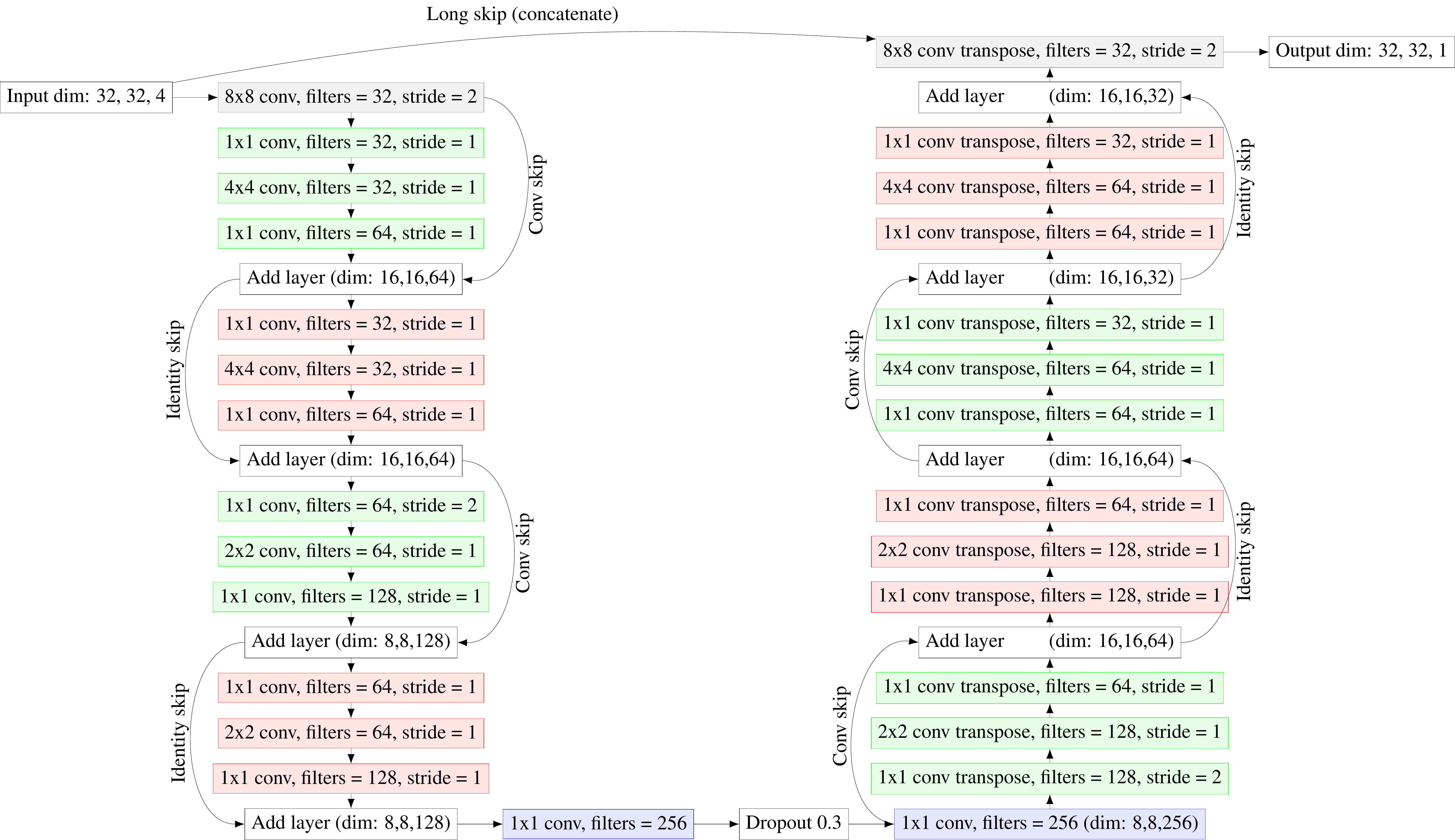}
    \caption[Encoder-Decoder ResNet as predictor]{Architecture of the developed fully convolutional neural network. The encoder is on the left side, the decoder on the right side and in between is the latent space. Identity- and convolutional-blocks are alternated and each block has a residual shortcut.}
    \label{fig:EncDec_ResNet}
\end{figure*}

To account for the locality of the cell transitions and because we are working with image data, we use convolutional layers. The layers are organized in several building blocks, with each block consisting of three convolutional layers. The first and last layer of each convolutional block have a kernel size of $1 \times 1$, the middle layer of either $4 \times 4$ or $2 \times 2$. At the beginning and at the end we use an $8 \times 8$ sized kernel. To provide regularization and reduce overfitting, we use Batch Normalization layers \cite{Ioffe2015} after each convolutional layer and use a Dropout layer in the latent space. The structure of the decoder is similar to the encoder but using deconvolution layers (the inverse of convolution), see figure \ref{fig:EncDec_ResNet}. For the optimization, the Adam optimizer \cite{Kingma2017} with a learning rate of $\eta = 0.0001$ was selected and the network was trained for $5000$ epochs.

\textit{Short range residual skip connections:} As we are stacking several convolutional layers on top of each other, we also make use of the short range skip connections from the ResNet. This is to ensure that network benefits from improved gradient flow due to residual design. To accommodate skip connections between layers with different dimensions, we use not only identity skip connections ($\mathcal{F}(x) + x = \mathcal{H}(x)$), but also convolutional skip connections ($\mathcal{F}(x) + conv(x) = \mathcal{H}(x)$). This provides an easy way to avoid using techniques like interpolation or zero padding. We also use one concatenative long range skip connection from the input to the output layer of the decoder.

\section{Results and Evaluation}

\begin{figure}
    \includegraphics[width = \textwidth]{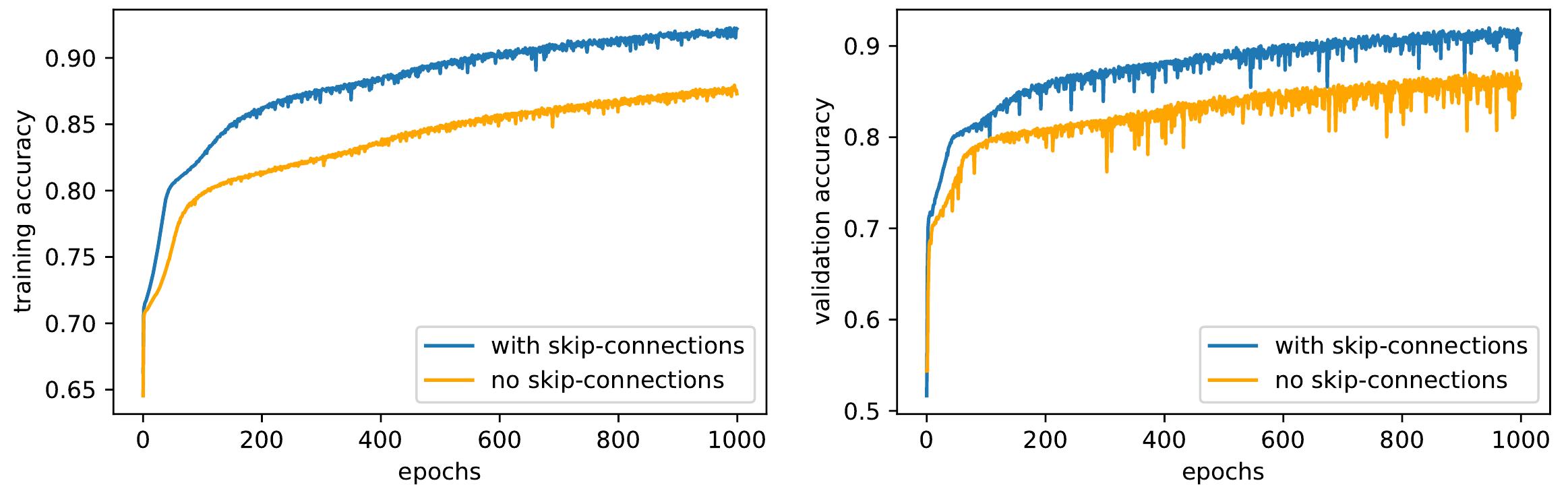}
    \caption[The effect of Skip Connections]{Comparison of training and validation results (for Level 1 generalization data) when enabling and disabling long and short range residual skip connections.}
    \label{fig_comparison_skip}
\end{figure}

\paragraph{Long range and short range residual skip connections:} 
Comparing the performance of a neural network with a concatenative long range skip connection and additive short range residual skip connections and neural network without any skip connections , we can see in figure \ref{fig_comparison_skip} that they play an important role in reducing the CA state prediction error. While they behave similarly during the first hundred epochs, the no-skip network suffers from the degradation problem later on. Here the skip connections can show their strength in further reducing the error both in the training and validation set. 

\paragraph{Simple generalization:}
As expected we do see very good results (see table \ref{results_gen}) when it comes to these tasks, with the accuracy averaging over $98\%$ across all datasets. The network seems to be very capable of learning the shown rules with some minor errors. For a Moore neighborhood of $3$ there are $ 3\times 3 -1 = 8$ cells that have two possible states, so we count $2^8 = 512$ possible cell transition rules. This "low" number can apparently easily be learned. Therefore we can take the calculated results as an upper boundary and best-case reference for all other levels of generalization. 

\paragraph{Level 1 generalization:}
For the Level 1 generalization, we can see that the overall accuracy decreases, as the network has to adapt to a greater number of diverse transition rules. However, the results are still very good with an average accuracy of $92-93\%$ over all data sets. This shows the ability of the network to learn a substantial number of the given rules, and apply them to new initial configurations, even for different rule neighborhood sizes. The number of possible transition rules becomes incredibly large ($ \approx 2^{(5 \times 5)}$ and $2^{(7 \times 7)}$), so the drop in accuracy compared to the simple generalization makes sense. Evaluating which neighborhood sizes influence the error the most, we split the validation and test data by neighborhood sizes again: about $98\%$ accuracy is achieved on the $3 \times 3$ data, $93\%$ on the $5 \times 5$ data, and $88\%$ on the $7 \times 7$ data. This also supports the argument that the larger the neighborhood size, the harder it gets to predict. 

\paragraph{Level 2 generalization:}
While the results for the training and validation dataset stay the same, we can see that the test loss and accuracy deteriorate down to $77\%$. Apparently the network, although being shown a diverse set of rules of a similar kind, is not able to adapt very well to completely unseen rules. However the results for the test set are still a lot better than the accuracy of the baseline, so it does provide some advantage. 

\paragraph{Level 3 generalization:}
For the extrapolation task we observe a similar effect as for the level 2 generalization, with a huge drop in accuracy for the test set. The loss is even larger than in the previous generalization step because the rules in the test set now are not only completely new but also based on a different neighborhood size. While still a lot better than the baseline accuracy, this task becomes apparently too hard for the network. It is however interesting to observe, as the results for the level 3 interpolation task show, we can increase the test outcome by $7 \% $ if we train on neighborhood sizes $9 \times 9$, that are larger than the test neighborhood size $7 \times 7$. As some cell transition rules in the $9 \times 9$ dataset will probably be very similar to the ones in the $7 \times 7$ one, the network is able to interpolate them. 

\begin{table}[!tb]
\centering
\begin{tabular}{c|c|c|c}
                           & train accuracy & val accuracy & test accuracy \\
                           \hline
Simple generalization      & $98,74 \pm 1,15$     & $98,44 \pm 1,20$   & $98,49 \pm 1,09$    \\
Level 1 generalization     & $93,00 \pm 1,00$   & $92,16 \pm 1,03$   & $92,19 \pm 1,12$    \\
Level 2 generalization     & $93,65 \pm 0,40$     & $92,76 \pm 0,47$   & $77,65 \pm 1,32$    \\
Level 3 extrapolation     & $92,63 \pm 1,35$     & $91,50 \pm 1,20$   & $70,50 \pm 1,77$    \\
Level 3 interpolation & $91,16 \pm 1,03$     & $90,40 \pm 0,83$   & $77,49 \pm 4,31$    
\end{tabular}
\caption{Results (accuracy in \%) for the different levels of generalization averaged over 5 runs. Baseline accuracy is $\approx 51 \%$. }
\label{results_gen}
\end{table}

 \bigskip

While a qualitative analysis of the results is hard to perform since the underlying patterns are hard to spot with the naked eye, we want to however report some interesting observations we made when looking at the rules where the network had the highest accuracy: It seems that "static" rules, that do not exhibit many changes in between time steps are very easy to predict. It makes sense that such a rule is very easy to extract for the neural network, as it can just copy the last time step of the data and only make minor modifications to it. Another set of rules that the model can successfully predict is what we call "alternating" rules. This kind of functions just switch the state of all cells from iteration to iteration to their opposite, as all alive cells die from overpopulation and all dead cells are reborn, see figure \ref{fig:trajectory}.  In this case, it is also clear that the error rate is so low because taking the last iteration and just flipping all the states constitutes a simple operation. Still, for the given rules the alternating behavior is not valid for all cells and when taking a closer look at the pictures, we can spot multiple non-alternating cells, that the network also predicted correctly. Following the classification system introduced by Wolfram \cite{Wolfram2002}, the static rules would correspond to \textit{Class I} and the alternating rules to \textit{Class II} rules.

\section{Discussion \& Conclusion}
The combination of several convolutional layers following the structure of the ResNet and U-Net \cite{He2016,Ronneberger2015} provides a viable  architecture for learning to predict  CA state evolution  from trajectories of states generated by CA's transition rules. The prediction results show the network is able to learn from a diverse range of these simple rules and generalize to unseen initial configurations successfully with an accuracy of $> 90 \%$. However, the more complex the rules (measured by the neighborhood size), the harder it becomes for the network to learn them. Performing on completely unseen rules or unseen neighborhood sizes in the test set is also hard, but still yields an accuracy of $>70 \%$, being significantly higher than the random guess baseline. 

Overall the results show that using a deep feed-forward network that exploits the local interactions of CA with convolutional layers using fixed-size kernels is sufficient to predict the behavior of a large range of CA with rules of different neighborhood sizes. As we only used the square-shaped Moore neighborhood in our study, several new levels of generalization could be created by experimenting with other shaped neighborhoods. By picking arbitrary cells anywhere on the grid as neighbors one could determine how well a network can also extract global rules instead of just local ones. It is worth noting that in terms of network structure the training did not converge once we placed a dense layer in the latent space, so the fully convolutional structure which emphasizes spatially local operations seems to play important role here. 

A recurrent or large transformer~\cite{Vaswani2017, Lu2021} network architecture may further increase the performance as the network would be able to infer the rules over a longer sequence of observations and eventually correct for mistakes done by a purely feed-forward convolutional network, which may also enable a deep learning model with stronger generalization as compared to the level observed in the current study. Recurrent or transformer architectures may also expand tractable time horizon for predicting multiple CA state steps. 

For further work, it would be also interesting to examine if it is possible to extract the rules directly from the latent space in explicit form. Here we could benefit from the simplicity of discrete CA rules again, as they are very easy to represent, e.g. with the Born/Stay notation. In this case, the output of the neural network would have to be just a set of integer numbers. This would also allow for further analysis on which specific rules the network fails to predict correctly and why, opening further venues towards systematic design of explainable AI architectures. 

On a broader scope, CA models could be used not only for testing the generalization ability of neural networks but might also serve as models or data generating simulators of the real world, given the Turing completeness of some CA (like the GoL \cite{Berlekamp2004}) and the ability to generate global complex behavior based on simple local rules. As stated by multiple works before, real world physics may be well captured by rather simple CA models \cite{Zuse91, Wolfram2002, wolfram2020project}. Especially the kind of complex self-organization processes that arise in biology and cannot be well explained by classical mathematical models that rely on global variables and their interactions would be a good candidate for CA models. CA are capable of generating self-reproducing patterns from local variables and their interactions and therefore fulfill a crucial requirement for modelling biological and artificial life. Some works are already beginning to exploit the potential of CA and learning neural networks to infer local rules necessary for building up complex global structures~\cite{Mordvintsev2020, Hernandez2021} or to discover rules leading to self-organizing patterns with complex, stable dynamics ~\cite{Reinke2019, Etcheverry2020}  

Seeing CA as data generator able to model complex processes via compact, simple local rules, a neural network that can understand the behavior of a CA by observation and extract the underlying latent rules would also be able to do the same with any given natural phenomena. CA can thus provide a very important tool to systematically test ability of neural networks and other learning algorithms to learn latent laws that govern complex phenomena from the raw observations, with a crucial ingredient of CA rules being fully accessible as ground truth behind the generated complex dynamics observed by the candidate learning algorithm.


%
%

\bibliographystyle{unsrt}
\bibliography{bibliography}   

%
%

\end{document}